\pdfoutput=1

\documentclass[11pt]{article}

\usepackage[final]{acl}

\usepackage{times}
\usepackage{latexsym}

\usepackage[T1]{fontenc}

\usepackage[utf8]{inputenc}

\usepackage{microtype}

\usepackage{inconsolata}

\usepackage{graphicx}

\usepackage{booktabs}
\usepackage{tabularx}
\usepackage{subcaption}
\usepackage{tcolorbox}  
\usepackage{lipsum} 

\usepackage{xcolor}
\usepackage{pifont} 
\usepackage{tikz} 
\usepackage{soul} 
\usepackage{multirow}
\usepackage{dblfloatfix}

\newcommand{\more}[1]{\textcolor{teal!80}{\tiny{#1}}}
\newcommand{\less}[1]{\textcolor{red}{\tiny{#1}}}

\title{Revisiting 3D LLM Benchmarks: Are We Really Testing 3D Capabilities?}

\author{
 \textbf{Jiahe Jin\textsuperscript{1*}}, 
 \textbf{Yanheng He\textsuperscript{1*}}, 
 \textbf{Mingyan Yang\textsuperscript{1*}} \\
 \textsuperscript{1}Shanghai Jiao Tong University
}

\usepackage{titling}

\begin{document}
\maketitle

\renewcommand{\thefootnote}{\fnsymbol{footnote}} 

\footnotetext[1]{Equal Contribution.}

\begin{abstract}
In this work, we identify the ``2D-Cheating'' problem in 3D LLM evaluation, where these tasks might be easily solved by VLMs with rendered images of point clouds, exposing ineffective evaluation of 3D LLMs' unique 3D capabilities. We test VLM performance across multiple 3D LLM benchmarks and, using this as a reference, propose principles for better assessing genuine 3D understanding. We also advocate explicitly separating 3D abilities from 1D or 2D aspects when evaluating 3D LLMs. Code and data are available at \href{https://github.com/LLM-class-group/Revisiting-3D-LLM-Benchmarks}{Github}.

\end{abstract}
\section{Introduction}

Recent advances in Large Language Models~\cite{chatgpt,gpt4} have led to the development of Vision Language Models (VLMs)~\cite{llava,gpt4o}. To overcome their lack of grounding in the real 3D physical world~\cite{3dllm}, researchers developed 3D LLMs~\cite{shapellm, pointllm} for 3D processing.

Given the extreme scarcity of 3D training data~\cite{3dllm, llava3d}, many approaches leverage existing LLMs and VLMs to generate annotations for 3D data~\cite{3dllm, gpt4point, pointllm}. This reliance on 2D and language (1D) priors raises a fundamental question: \textit{What capabilities do 3D LLMs possess that truly differentiate them from 2D VLMs?}

\begin{figure}[h]
  \includegraphics[width=\columnwidth]{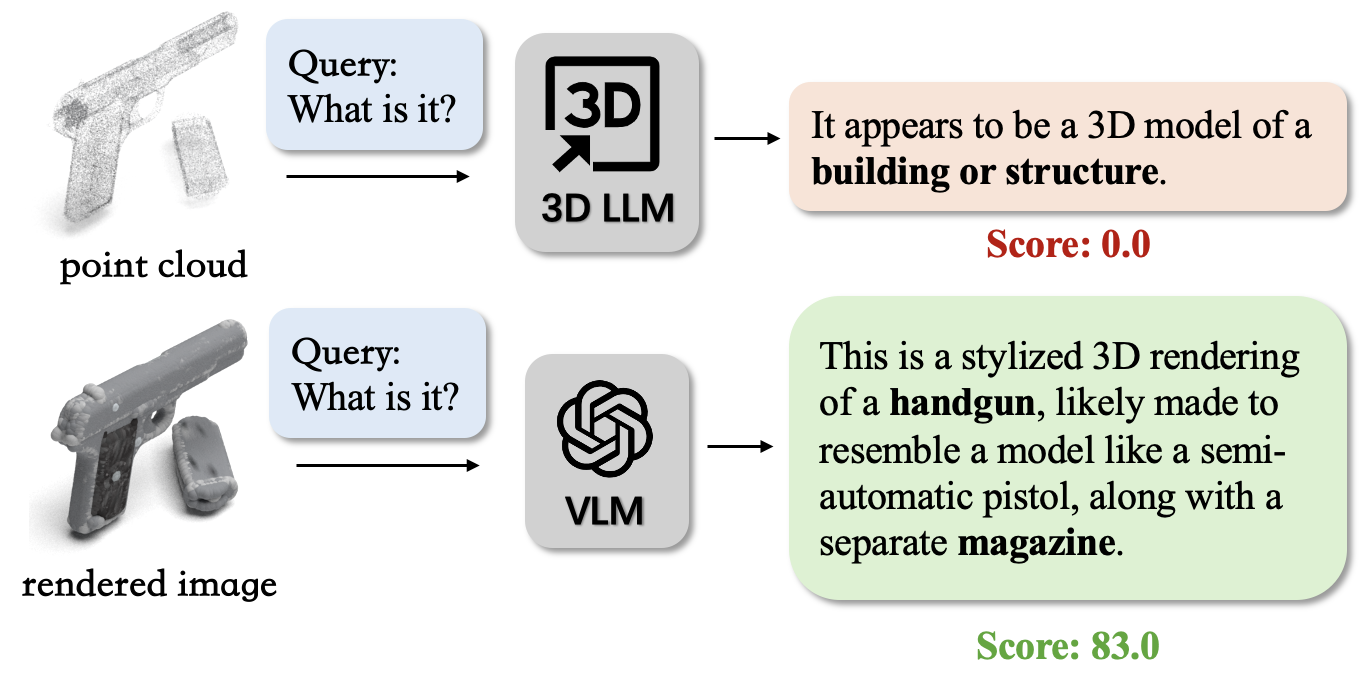}
  \caption{Example of 2D-Cheating. With rendered images of the point cloud, VLMs could easily solve some 3D tasks, and even outperform 3D LLMs.}
  \label{fig:2d cheating}
\end{figure}

To explore this, we revisit some benchmarks used for 3D LLM evaluation. Current 3D LLMs are primarily evaluated on Q\&A or captioning tasks~\cite{3dllmsurvey}, rather than specific downstream tasks like object detection~\cite{nuscenes,fcos3d}, as LLMs provide a general-purpose language interface~\cite{lminterface}. As shown in Figure~\ref{fig:2d cheating}, we find some tasks can be easily solved by 2D VLMs, indicating that they fail to effectively assess the unique capabilities of 3D models. We refer to this phenomenon as \textbf{2D-Cheating}.

Specifically, we render point clouds into images to test the performance of VLMs on 3D LLM benchmarks. Experimental results indicate that \textbf{VLMs can significantly exceed the state-of-the-art (SOTA) performance of 3D models on certain benchmarks}. However, on other benchmarks, even with enhanced information, they still fail to achieve the best performance.

We argue that tasks where VLMs perform poorly compared to 3D LLMs are those that truly involve unique 3D capabilities. Based on analysis, we (1) propose several principles for designing benchmarks that effectively evaluate true 3D capabilities, and (2) advocate for separating the evaluation of 3D capabilities from 2D and 1D aspects when assessing 3D LLMs.

\vspace{-6pt} 

\section{Method}

\subsection{VLM3D}

\begin{figure}[h]
  \includegraphics[width=\columnwidth]{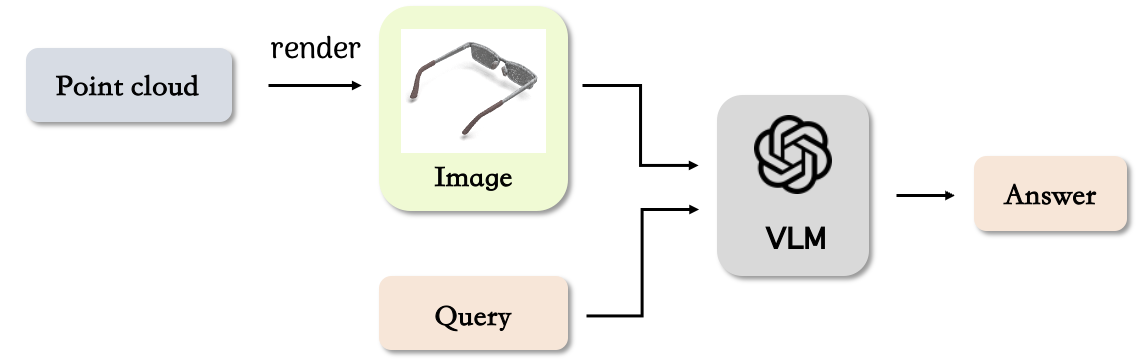}
  \caption{The VLM3D pipeline. For a given task, we render the point cloud into images and feed both the images and the query into a VLM.}
  \label{fig:VLM3D}
\end{figure}

As shown in Figure~\ref{fig:VLM3D}, we propose \textbf{VLM3D}, a simple yet general pipeline that adapts VLMs to 3D tasks. Specifically, it first renders point clouds into images and augments queries with few-shot, then feeds them into a VLM.

\subsection{Viewpoint Selection}
\label{sec:view}

Viewpoint of image rendering significantly impacts VLM's input information about the 3D asset. We hypothesize that a key limitation of 2D models in 3D understanding stems from the \textbf{viewpoint-specific nature of images}, whereas point clouds inherently provide a holistic 3D representation. The challenges posed by viewpoint dependency include: (1) blind spots in areas outside the selected viewpoint; (2) occlusion and overlap of objects; and (3) single-surface capture that lacks multifaceted geometry. Based on this, we set different viewpoint rendering configurations for object and scene point cloud benchmarks.

\paragraph{Object}
Object point clouds are relatively simple, so we use a fixed viewpoint for each benchmark. Following common practice, the camera is positioned at a fixed location looking toward the origin of the point cloud. No specific adjustments are made for individual point clouds.

\paragraph{Scene}
Scene point clouds present greater complexity. To systematically investigate viewpoint impacts, we conduct the following experimental settings, as all viewpoints illustrated in Figure~\ref{fig:multi-view}.

\subsubsection{Single View}
\label{sec:single view}
In this setting, we follow the common practice~\cite{sqa3d} to render images from Bird's Eye View (BEV).

\subsubsection{Multi View}
\label{sec:multi view}
In this setting, we render images from four fixed viewpoints (East, South, West, and North) and combine them into a multi-view image.

\subsubsection{Oracle View}
\label{sec:oracle view}
\begin{figure}[h]
  \centering
  \includegraphics[width=0.4\textwidth]{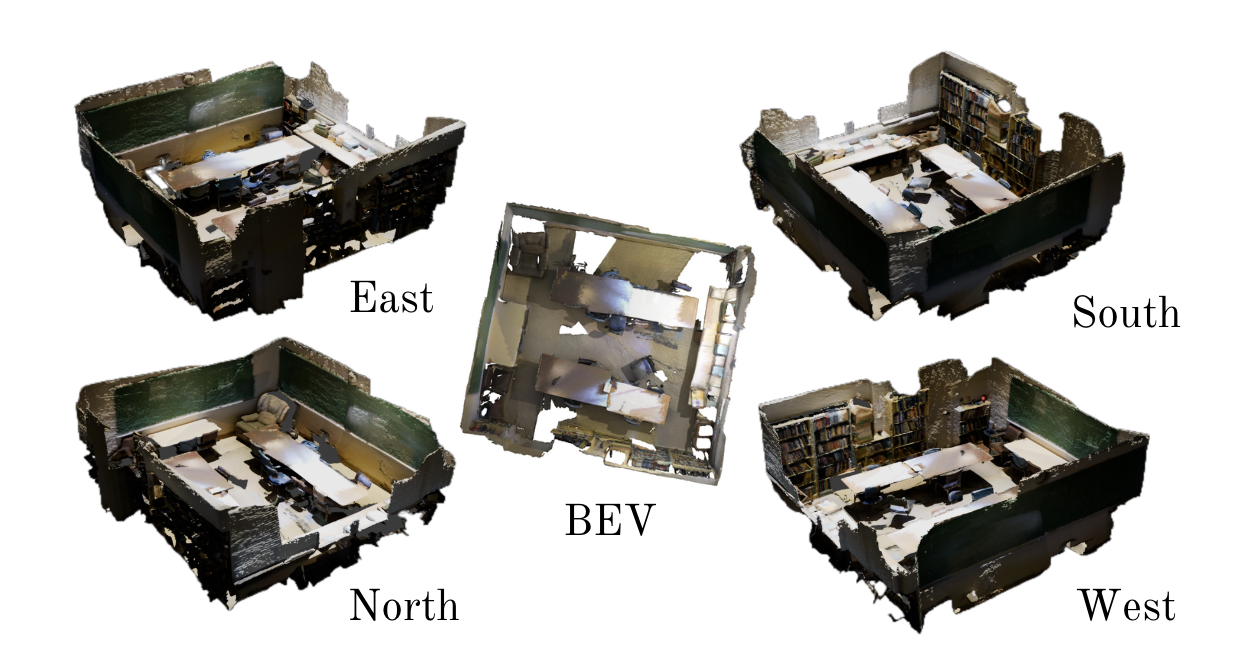}
  \caption{All five viewpoints of an example scene.\newline}
  \label{fig:multi-view}
\end{figure}

In this setting, we conduct experiments on the validation set to explore the upper limits of VLM3D’s capabilities. We first render images from all five viewpoints shown in Figure \ref{fig:multi-view}, and then use the \textbf{Best-of-N} (BoN) method to select the optimal view for each question.

Since LLMs are probabilistic, simply taking the highest evaluation score risks inflating the likelihood of randomly guessing the correct answer. To address this, we sample multiple responses for each viewpoint and calculate the average score. The highest average score among all viewpoints is then selected as the oracle view score.

\begin{figure*}[h]
    \centering
    \includegraphics[width=1\linewidth]{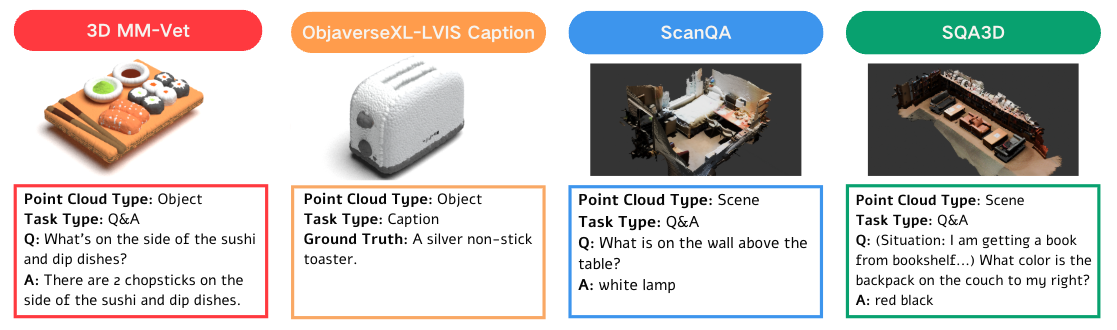}
    \caption{3D LLM benchmarks tested in our experiments, see detailed introduction in Appendix~\ref{sec:benchmarks}.}
    \label{fig:benchmark overview}
\end{figure*}

\subsection{Other Rendering Factors}
While other factors like rendering style may affect image quality, viewpoint selection remains the trickiest due to the difficulty of a unified optimal setup. For each benchmark, we apply a consistent set of other rendering configurations across all point clouds to ensure high-quality results.

\section{Experiments}

\subsection{Setup}
We conduct experiments on the benchmarks in Figure~\ref{fig:benchmark overview}, covering major benchmarks used for evaluating 3D LLMs. We compare VLM3D's performance with current SOTA 3D LLMs: ShapeLLM-13B on 3D MM-Vet~\cite{shapellm}, GPT4Point on ObjaverseXL-LVIS Caption~\cite{objaverse}, LEO~\cite{leo} on ScanQA~\cite{sqa3d}, and SIG3D~\cite{sig3d} on SQA3D~\cite{sqa3d}. For scene point cloud benchmarks, we also compare against the best-performing model at the time of the benchmark’s release as 3D Baseline. For the VLM used in VLM3D, we choose GPT-4o~\cite{gpt4o} and Qwen2-VL-72B~\cite{qwen2vl}.

\subsection{Object Point Cloud Benchmark}
For object point cloud benchmarks, we choose 3D MM-Vet for Q\&A and ObjaverseXL-LVIS Caption for captioning, with results presented in Table~\ref{tab:object}.

\begin{table}[h!]
\centering
\small
\setlength{\tabcolsep}{4pt}
\resizebox{0.5\textwidth}{!}{
\begin{tabular}{lccc}
\toprule 
\textbf{Benchmark} & \textbf{Metric}& \textbf{3D SOTA} & \textbf{VLM3D (GPT-4o)} \\
\midrule 
3D MM-Vet & LLM-eval & 43.2 & \textbf{58.1} \more{(+14.9)} \\
\midrule
\multirow{2}{*}{ObjaverseXL-} & BLEU-1 & 32.2 & \textbf{36.2} \more{(+4.0)} \\
\multirow{2}{*}{LVIS Caption} & ROUGE-L & 35.5 & \textbf{36.8} \more{(+1.3)} \\
& CIDEr & 78.0 & \textbf{79.3} \more{(+1.3)} \\
\bottomrule 
\end{tabular}
}
\caption{Results on object benchmarks.}
\label{tab:object}
\end{table}

\subsection{Scene Point Cloud Benchmark}
We conduct experiments on two widely used scene point cloud benchmarks, ScanQA and SQA3D. ScanQA evaluates general Q\&A, while SQA3D further assesses situation understanding.
\subsubsection{Single View Evaluation}
 We first conduct the single view evaluation mentioned in Section~\ref{sec:single view}, with results in Table \ref{tab:single view}.

\begin{table}[h!] 
\centering
\small
\setlength{\tabcolsep}{4pt}
\resizebox{0.5\textwidth}{!}{
\begin{tabular}{lccccc}
\toprule 
\multirow{2}{*}{\textbf{Benchmark}} & \multirow{2}{*}{\textbf{Metric}} & \textbf{VLM3D} & \textbf{VLM3D} & \multirow{2}{*}{\textbf{3D Baseline}} & \multirow{2}{*}{\textbf{3D SOTA}} \\
 & & \textbf{(Qwen2-VL)} & \textbf{(GPT-4o)} & & \\
\midrule 
& METEOR & 12.8 \less{(-0.3)} & 14.7 \more{(+1.6)} & 13.1 & 20.0 \more{(+6.9)}\\
\textbf{ScanQA(val)} & ROUGH & 27.9 \less{(-5.4)} & 25.8 \less{(-7.5)} & 33.3 & 49.2 \more{(+15.9)} \\
& CIDEr & 51.2 \less{(-13.7)} & 47.2 \less{(-17.7)} & 64.9 & 101.4 \more{(+36.5)}\\
\midrule 
\textbf{SQA3D(test)} & EM & 42.2 \less{(-5.0)}& 44.8 \less{(-2.4)}& 47.2 & 52.6 \more{(+5.4)} \\
\bottomrule 
\end{tabular}
}
\caption{Result for single-view evaluation on scene benchmarks. \textcolor{red}{Red} and \textcolor{teal!80}{green} indicate decrease and increase relative to the 3D Baseline, respectively.
}
\label{tab:single view}
\end{table}  

We observe that these benchmarks often focus on specific objects within intricate scenes, with single view images frequently missing required information. To further explore the upper bound of VLMs, we conduct experiments using different viewpoints on Qwen2-VL-72B.

\subsubsection{Multi View Evaluation}

We conduct multi-view evaluation based on the settings in Section~\ref{sec:multi view}, with results in Table~\ref{tab:multi-view}.

\begin{table}[h!]
\centering
\small
\setlength{\tabcolsep}{4pt}
\resizebox{0.5\textwidth}{!}{
\begin{tabular}{lccccc}
\toprule 
\textbf{Benchmark} & \textbf{Metric} & \textbf{SV} & \textbf{MV} & \textbf{3D Baseline} & \textbf{3D SOTA}\\
\midrule 
& METEOR & 12.8 \less{(-0.3)} & 13.1 \more{(+0.0)} & 13.1 & 20.0 \more{(+6.9)}\\
\textbf{ScanQA(val)} & ROUGH & 27.9 \less{(-5.4)}& 29.1 \less{(-4.2)} & 33.3 & 49.2 \more{(+15.9)} \\
& CIDEr & 51.2 \less{(-13.7)} & 54.0 \less{(-10.9)}& 64.9 & 101.4 \more{(+36.5)}\\
\midrule
\textbf{SQA3D(test)} & Overall & 42.2 \less{(-5.0)} & 43.2 \less{(-4.0)} & 47.2 & 52.6 \more{(+5.4)} \\
\bottomrule 
\end{tabular}
}
\caption{The result of multi-view evaluation on scene benchmarks. SV refers to single-view, and MV refers to multi-view.}
\label{tab:multi-view}
\end{table}

\subsubsection{Oracle View Evaluation}
\begin{figure}[h!]
    \centering
    \includegraphics[width=0.5\textwidth]{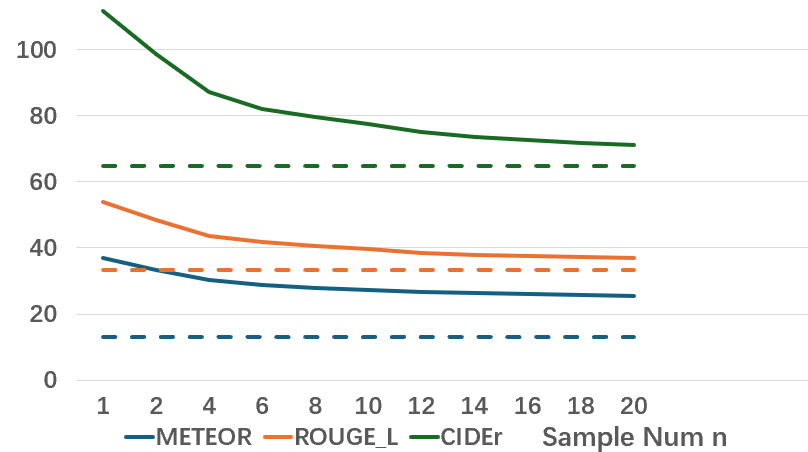}
    \caption{Result of BoN on ScanQA(val) with $k=5$. The dashed line represents the 3D baseline ScanQA, and the solid line shows VLM3D’s performance.\newline}
    \label{fig:best-of-n}
\end{figure}

We perform oracle view evaluation using the \textbf{Best-of-N} method in Section~\ref{sec:oracle view}. For each question, we render images from $k$ viewpoints and sampled $n$ responses per viewpoint. As shown in Figure~\ref{fig:best-of-n}, the result decreases as $n$ increases and converges around $n=20$, indicating that randomness from selecting the maximum score is effectively mitigated. The scores are detailed in Table~\ref{tab: BoN}.

Besides the Best-of-N, we explore another method called \textbf{Human-Intuition-Selection} (HIS). We calculate the centroids of all relevant objects of each question and use a heuristic algorithm to select the best viewpoint that captures the centroids effectively. Details and experimental results of these method are in Appendix~\ref{sec:human-intuition-selection}.

\setulcolor{red!70}
\begin{table}[h!]
\centering
\small
\setlength{\tabcolsep}{4pt}
\resizebox{0.49\textwidth}{!}{
\begin{tabular}{lcccccc}
\toprule 
\textbf{Metric} & \textbf{SV} & \textbf{BoN(k=4)}& \textbf{BoN(k=5)}& \textbf{3D Baseline}& \textbf{3D SOTA}\\
\midrule 
METEOR & 12.8 \less{(-0.3)} & 27.0 \more{(+13.9)} & 28.2 \more{(+15.1)} & 13.1 & 20.0 \more{(+6.9)}\\
ROUGH & 27.9 \less{(-5.4)} & 35.6 \more{(+2.3)} & 36.9 \more{(+3.6)} & 33.3 & 49.2 \more{(+15.9)}\\
CIDEr & 51.2 \less{(-13.7)} & 68.4 \more{(+3.5)} & 71.2 \more{(+16.3)} & 64.9 & 101.4 \more{(+36.5)}\\
\bottomrule 
\end{tabular}
}
\caption{Result of oracle view on ScanQA(val). the sample number of BoN is $n=20$.}
\label{tab: BoN}
\end{table} 

\begin{figure*}[h!]
    \centering
    \includegraphics[width=0.9\linewidth]{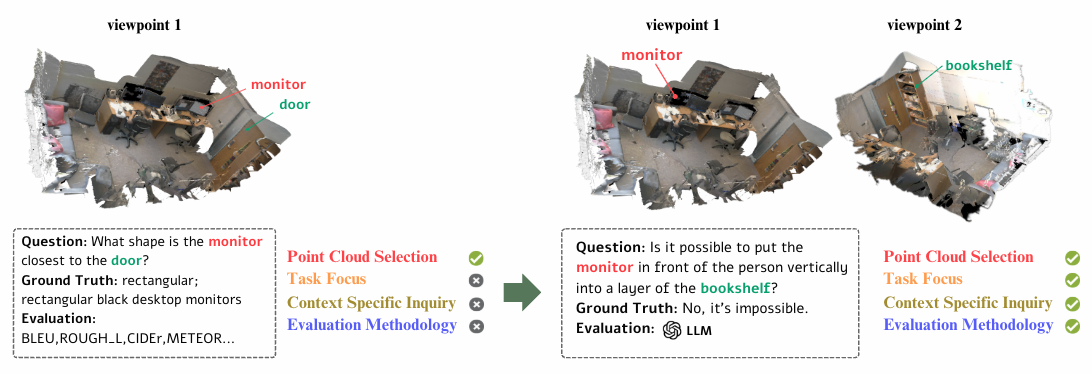}
    \caption{An example of task redesign, with the original task on the left and the new task on the right.}
    \label{fig: redesign example}
\end{figure*}

\subsection{Human Evaluation}
\label{sec: human evaluation}
We performed human evaluation on 50 randomly sampled tasks from each benchmark, and use Spearman's rank correlation (rho)~\cite{spearman1904association} to calculate the correlation between original metrics and human judgments, as shown in Table~\ref{tab:rho}.


\begin{table}[h!]
\centering
\small
\setlength{\tabcolsep}{4pt}
\resizebox{0.33\textwidth}{!}{
\begin{tabular}{lcc}
\toprule 
\textbf{Benchmark} & \textbf{Metric} & \textbf{rho} \\
\midrule 
\textbf{3D MM-Vet} & LLM evaluation & 77.7 \\
\midrule
\multirow{2}{*}{\textbf{ObjaverseXL-}} & Bleu\_1 & 49.8 \\
\multirow{2}{*}{\textbf{LVIS Caption}} & ROUGH\_L & 60.1 \\
 & CIDEr & 50.9 \\
\midrule
& METEOR & 28.3 \\
\textbf{ScanQA} & ROUGH\_L & 33.6 \\
& CIDEr & 25.5 \\
\midrule
\textbf{SQA3D} & EM & 80.2 \\
\bottomrule 
\end{tabular}
}
\caption{The rho between original benchmark metrics and human judgments.}
\label{tab:rho}
\end{table}
\section{Analysis}

\subsection{Object Point Cloud}
Results demonstrate that VLM significantly outperforms SOTA 3D LLMs using only single view images on these object benchmarks. This suggests that these tasks can be easily solved without specialized 3D representations, indicating a lack of effective 3D capability assessment. The limitation may arise from (1) \textcolor{red!70}{\LARGE $\bullet$} \ul{the inherent simplicity of these object point clouds}, 
and (2) \textcolor{orange!70}{\LARGE $\bullet$}\label{point1}\setulcolor{orange!70} \ul{these tasks primarily require only surface-level information that could be equally extracted from a simple rendered image.} See more task examples in Appendix~\ref{sec:3d-mm-vet} and~\ref{sec:obj-caption}. 

\subsection{Scene Point Cloud}
\subsubsection{Poor Performance with Single View}

\setulcolor{red!70}
In these scene benchmarks, even the most advanced VLMs consistently underperform top 3D models with the single view setting, \textcolor{red!70}{\LARGE $\bullet$} \ul{revealing their limitations in handling complex 3D scenes}.

\setulcolor{blue!60}
Besides, \textcolor{blue!60}{\LARGE $\bullet$} \ul{many benchmarks rely on text similarity metrics for evaluation, which is overly rigid given the flexible nature of natural language}. See case study in Appendix~\ref{sec: eval metric}.

\setulcolor{orange!70}
\subsubsection{Challenges of Multi View Understanding}
As shown in Table~\ref{tab:multi-view} \&~\ref{tab: BoN}, providing multi-view images only brings a slight improvement. In contrast, oracle view leads to a significant boost. Notably, when setting $k=4$, all candidate viewpoints of Best-of-N are included in the multi-view inputs, theoretically providing sufficient information. However, \textcolor{orange!70}{\LARGE $\bullet$} \ul{results indicate that VLMs struggle to form a unified understanding of a scene from multi-view}.

\subsubsection{Enhancement with Oracle View}
As shown in Table~\ref{tab: BoN}, oracle view performance significantly surpasses the 3D Baseline, confirming that providing good viewpoints greatly enhances VLM’s potential to solve these tasks through 2D-Cheating. However, dynamically providing the oracle viewpoint for each question considerably simplifies the task by identifying and presenting the key information from the complete scene.

Besides, the performance of BoN still lags behind the 3D SOTA. This suggests that, \textcolor{orange!70}{\LARGE $\bullet$} \ul{even provided with a favorable view, VLMs still struggle to match the best performance of 3D LLMs in some tasks}.

\subsubsection{Failure of HIS Method}
Table~\ref{tab:two-oracle-view} shows that HIS performs significantly worse than Best-of-N. Through case studies, we identify three key factors: (1) \textcolor{orange!70}{\LARGE $\bullet$} \ul{Selecting the best viewpoint for detail items in complex scenes with intuition is challenging}, and the resulting viewpoints are sometimes suboptimal, primarily due to object occlusions. \setulcolor{blue!60}(2) \textcolor{blue!60}{\LARGE $\bullet$} \ul{Limited ground truth reveals a clear preference}. As detailed in Appendix~\ref{sec: ground truth prefer scanqa}, while the viewpoints selected by the HIS method do provide useful information, BoN selects viewpoints that align most closely with the ground truth by definition. As a result, although answers from the HIS method are reasonable, the limited coverage of ground truth makes BoN methods achieve higher scores in some cases. (3) \textcolor{olive!90}{\LARGE $\bullet$} \setulcolor{olive!90} \ul{Many questions rely on common sense, not the specific details of the point cloud}, as shown in Appendix~\ref{sec: general knowledge}. Specific information in HIS viewpoints might even lead to a worse score, compared to answering leveraging world knowledge with an unrelated viewpoint in BoN.

\section{Principles For Effective 3D Evaluation}
Based on the analysis, we propose the following principles for effective 3D capabilities evaluation of 3D LLMs.\footnote{The analyses leading to each principle are marked with corresponding colored circular markers in the preceding text.} We also apply these principles to redesign a task, as shown in Figure~\ref{fig: redesign example}.

\subsection{Principles}
\paragraph{\textcolor{red!70}{\LARGE $\bullet$} \textcolor{red!70}{Point Cloud Selection}}
A complex point cloud should be selected, such as a point cloud of scenes or objects with more intricate structures.

\paragraph{\textcolor{orange!70}{\LARGE $\bullet$} \textcolor{orange!70}{Task Focus}} Tasks should go beyond general surface-level information, diving into the intricate details of 3D assets. This includes the structural specifics of individual objects and the finer items within complex scenes. Furthermore, questions should prioritize aspects that cannot be easily answered with images from any viewpoints.

\paragraph{\textcolor{olive!90}{\LARGE $\bullet$} \textcolor{olive!90}{Context Specific Inquiry}} Avoid overly general questions, ask the unique aspects of the current 3D asset, and try to ``violate common sense'' to test genuine understanding of 3D input.

\paragraph{\textcolor{blue!60}{\LARGE $\bullet$} \textcolor{blue!60}{Evaluation Methodology}} For problems with multiple reasonable answers, include various possible answers to accommodate more reasonable results and varying detail levels. Besides, avoid text similarity metrics and instead rely on LLMs for more flexible assessments.

\subsection{Quantitative Validation}

We conduct a quantitative analysis to validate the effectiveness of our principles. The results demonstrate that \textbf{existing benchmarks indeed fall short of these principles to varying degrees}.

\paragraph{LLM Evaluation}

We conduct LLM evaluation on the first three principles related to task design. Specifically, we provide GPT-4o with all the queries, ground truth, and rendered point clouds and ask it to determine whether each task complies with these principles.

\begin{table}[h!]
\centering
\small
\setlength{\tabcolsep}{4pt}
\resizebox{0.5\textwidth}{!}{
\begin{tabular}{lcccc}
\toprule
\multirow{2}{*}{\textbf{Pass Rate (\%)}} & \multirow{2}{*}{\textbf{3D MM-Vet}} & \textbf{ObjaverseXL} & \multirow{2}{*}{\textbf{ScanQA}} & \multirow{2}{*}{\textbf{SQA3D}} \\
 & & \textbf{-LVIS Caption} & & \\
\midrule
Point Cloud Selection & 53.9 & 36.3 & 69.7 & 86.2 \\
Task Focus & 25.9 & 19.0 & 76.5 & 91.6 \\
Context Specific Inquiry & 64.7 & 93.5 & 99.2 & 97.6 \\
\midrule
All Three Above & 16.0 & 7.3 & 62.9 & 81.9 \\
\bottomrule
\end{tabular}
}
\caption{The pass rate of each benchmark in LLM Evaluation.}
\label{tab:llm-evaluation}
\end{table}

As shown in the Table~\ref{tab:llm-evaluation}, even the benchmark with the highest pass rate (SQA3D) still has \textbf{18.1\%} of tasks failing to pass all three principles, and the varying pass rates across benchmarks correlate with the performance difference between VLMs and 3D LLMs. For example, lower pass rates are observed on object point cloud benchmarks where VLMs outperform 3D SOTA.

\paragraph{Human Evaluation}
Our human evaluation experiments in Section~\ref{sec: human evaluation} strongly support our last principle about evaluation methodology. Using LLM evaluation, 3D MM-Vet shows a high correlation with human judgments . SQA3D achieves an even higher correlation because it's a multiple-choice task. In contrast, open-ended tasks like ObjaverseXL-LVIS Caption and ScanQA, which rely on text similarity metrics, demonstrate lower correlations. These results support our advocacy for diverse possible answers and flexible LLM-based assessments for open-ended questions.

\section{Conclusion}
This paper identifies the 2D-Cheating problem in current 3D LLM evaluation, compares VLM performance with 3D LLMs across various benchmarks, and provides principles for assessing real 3D capabilities in 3D LLMs. However, this does not imply that 3D LLMs can ignore the ability to solve tasks vulnerable to 2D-Cheating. Therefore, we emphasize the need to deliberately decouple the evaluation of 3D capabilities from 1D and 2D capabilities to ensure that core 3D evaluations are not confused or overlooked.


\section{Limitations}
Our evaluation may not fully capture the capabilities of VLMs. First, the 3D LLMs were trained on the benchmark’s training set or similar tasks, whereas we only adapted VLMs to these benchmarks through few-shot prompting. Additionally, rendering point clouds into images inherently loses information compared to using real photographs. These factors could limit the assessment of VLMs’ real capability and may have weakened the observed 2D-Cheating issue.

Furthermore, our experiments were conducted solely on the Qwen2-VL and GPT-4o, which may not fully represent the performance of all available VLMs. The number of benchmarks we tested was limited, meaning the results might not generalize across a broader set of tasks. Ideally, a new benchmark that aligns with these principles would be developed, but due to constraints in space, resources, and other practical considerations, we leave this for future work.

\section{Ethics Statement}
This research on 3D LLM benchmarks adheres to ethical principles by exclusively utilizing publicly available data and maintaining transparency through open-source code. Our study does not involve any human annotation or privacy concerns. We do not anticipate any potential risks from this methodological study.

\section*{Acknowledgement}
We would like to express our sincere gratitude to Prof. Yonglu Li for his valuable guidance and support throughout this research, from topic selection to the final writing. His insightful discussions and feedback have been essential to the completion of this work.

\bibliography{custom}

\clearpage

\appendix

\section{Benchmarks in Experiments}
\label{sec:benchmarks}

To the best of our knowledge, given the nascent development of 3D LLMs, there is currently no dedicated and representative benchmark for their evaluation. We select representative benchmarks for 3D LLM evaluation: 3D MM-Vet and ObjaverseXL-LVIS Caption were proposed by model creators~\cite{shapellm, gpt4point}, while ScanQA and SQA3D are previously established 3D question-answering benchmarks~\cite{scanqa, sqa3d} that have been widely adopted in 3D LLM assessments.

\subsection{3D MM-Vet}
\label{sec:3d-mm-vet}

3D MM-Vet~\cite{shapellm} is a benchmark designed to evaluate multi-modal large language models in the context of 3D comprehension, particularly for embodied scenarios. The benchmark includes the following categories: \textbf{General Recognition}, \textbf{Knowledge}, \textbf{Language Generation}, \textbf{Spatial Awareness}, and \textbf{Embodied Interaction}. 

\begin{figure}[h!]
    \includegraphics[width=0.4\textwidth]{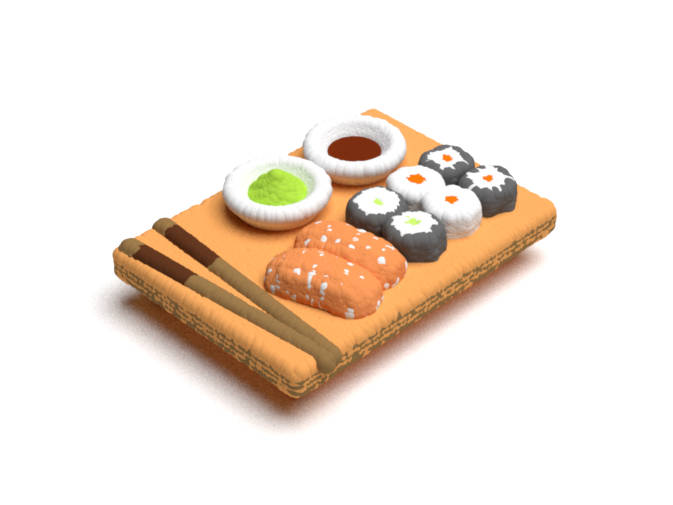}
    \caption{Example rendered image of a point cloud in 3D MM-Vet}
    \label{fig:3dmmvet_sample}
\end{figure}

Examples of each task type are shown below, using the same point cloud input depicted in Figure~\ref{fig:3dmmvet_sample}.

\begin{itemize}
    \item \textbf{General Visual Recognition}: \\
    Question: What is this?\\
    Answer: There's a board, on which are different kinds of sushi, one dish of Soy sauce, one dish of wasabi and chopsticks.
    \item \textbf{Knowledge}: \\
    Question: How many sushi will there be if 2 more sushi are added? \\
    Answer: There will be 8 sushi if 2 are added.
    \item \textbf{Language Generation}: \\
    Question: Describe its cuisine culture in detail. \\
    Answer: Sushi is a kind of Japanese cuisine, the main material is vinegar, sugar, salt seasoning, fat, and slightly sweet Japanese pearl rice cooked into vinegar rice, after cooling with fish, seafood, vegetables, eggs, or other meat and other ingredients.
    \item \textbf{Spatial Recognition}: \\
    Question: Which sushi is closer to the chopsticks, The white sushi or the black sushi? \\
    Answer: The black sushi is closer to the chopsticks.
    \item \textbf{Embodied Interaction}:\\
    Question: Give me steps to feed me eat this dish. \\
    Answer: Step 1: Pick up chopsticks. Step 2: Pick up the sushi with chopsticks. Step 3: Dip in soy sauce and mustard. Step 4: Move the chopsticks to get the sushi into your mouth.
\end{itemize}

\subsection{ObjaverseXL-LVIS Caption}
\label{sec:obj-caption}
GPT4Point~\cite{gpt4point} utilizes the Objaverse dataset, aligning it with LVIS categories \cite{LVIS}, to create the Objaverse-LVIS dataset. In this dataset, scenes featuring complex environments such as indoor houses or outdoor parks are excluded, focusing instead on scenarios involving individual objects or combinations of multiple objects.

\begin{figure}[htbp]
\centering
\begin{minipage}[t]{0.48\textwidth}
\centering
\includegraphics[width=0.45\textwidth]{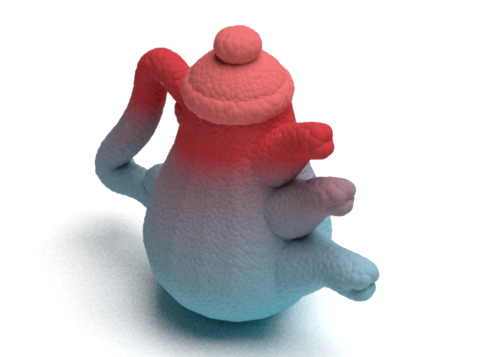}
\caption{Caption: A red and blue gradient teapot with three spouts.}
\label{fig:gpt4point_sample1}
\end{minipage}
\begin{minipage}[t]{0.48\textwidth}
\centering
\includegraphics[width=0.45\textwidth]{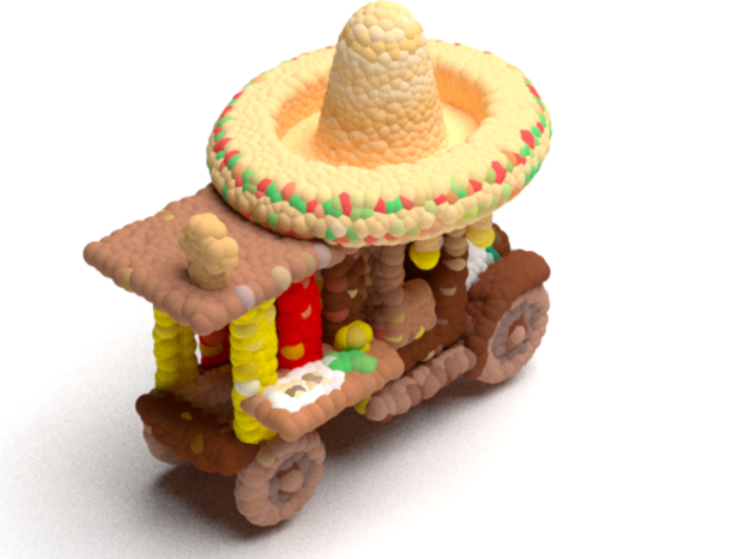}
\caption{Caption: A Mexican food cart decorated with traditional Mexican cuisine, topped with a sombrero hat.}
\label{fig:gpt4point_sample2}
\end{minipage}
\end{figure}

The ObjaverseXL-LVIS benchmark consists of two tasks: \textbf{3D object point cloud captioning} and \textbf{3D point cloud question answering}. Since only the captioning portion of the dataset is open-source, our experiments were conducted solely on this part. Figure \ref{fig:gpt4point_sample1} and \ref{fig:gpt4point_sample2} show two examples of captions.

\begin{figure*}[ht]
    \includegraphics[width=\linewidth]{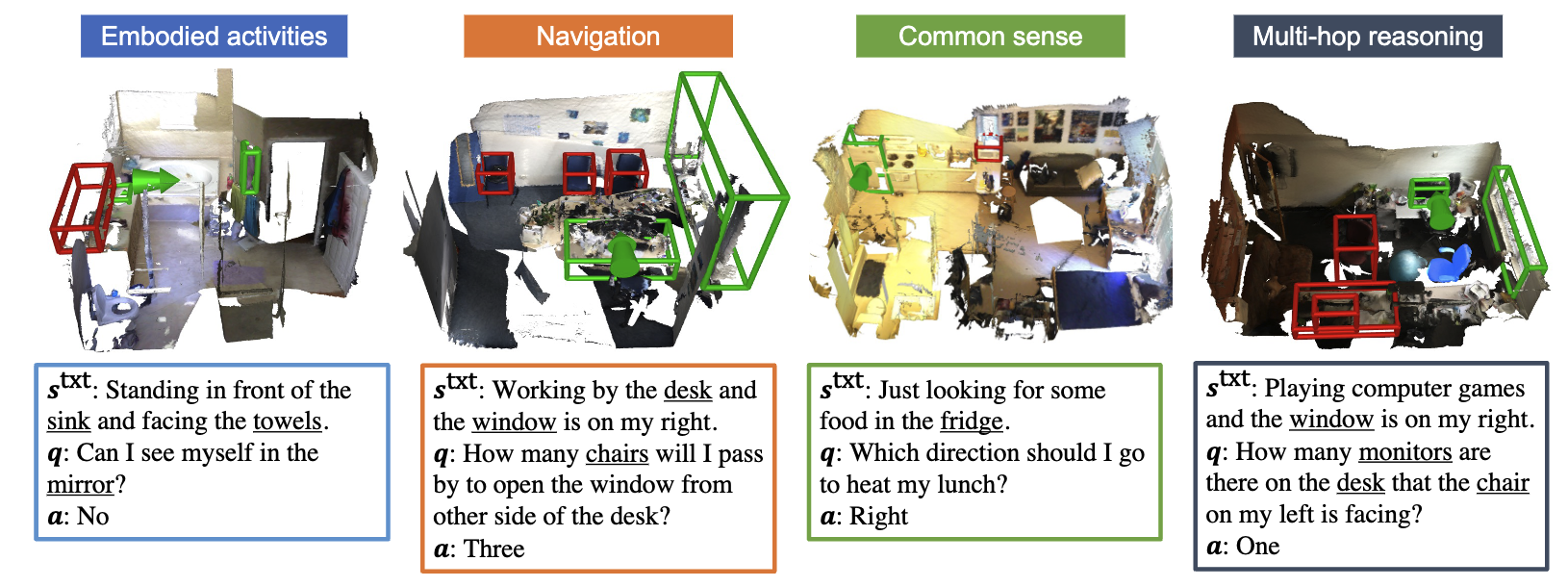}
    \caption{Examples from SQA3D.  
    \textit{Image from Ma et al., ``SQA3D: Situated Question Answering in 3D Scenes'', ICLR 2023, used under CC BY 4.0 license.}}
    \label{fig:sqa3dexample}
\end{figure*}

\subsection{ScanQA}
ScanQA~\cite{scanqa} was created for the 3D-QA task using RGB-D scans of indoor scenes and annotations from the ScanNet dataset. The dataset includes both question-answer pairs and 3D object localization annotations, making it one of the largest datasets for specifying object properties in 3D scenes via question-answering. The benchmark includes the following task categories: \textbf{Place}, \textbf{Number}, \textbf{Color}, \textbf{Object nature}, \textbf{Object} and \textbf{Other}. 

\begin{figure}[h!]
    \centering
    \includegraphics[width=0.4\textwidth]{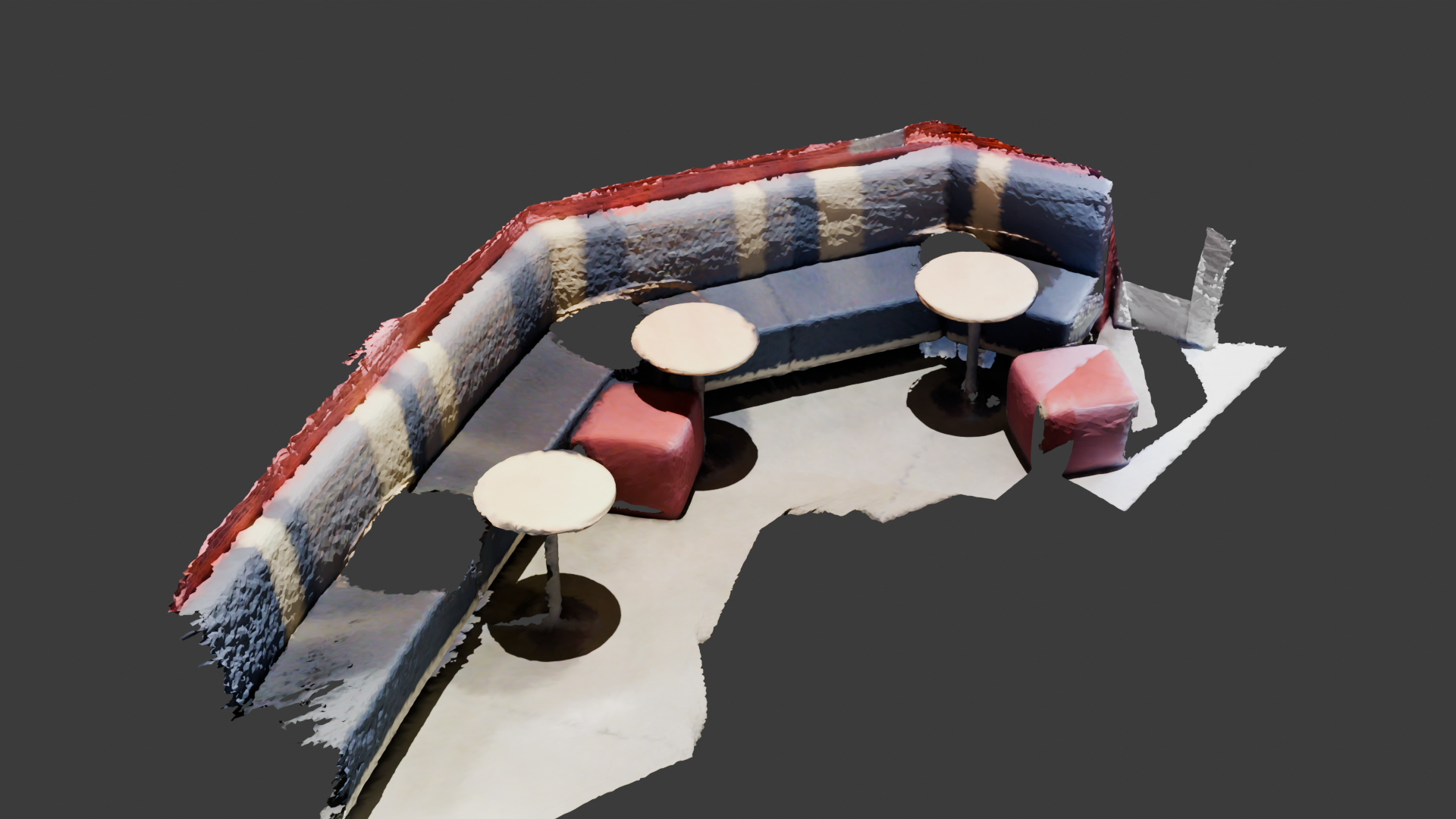}
    \caption{Example rendered image of a point cloud of the scene in ScanQA}
    \label{fig:scanqa_sample}
\end{figure}

Examples of each task type are shown below, using the same point cloud input depicted in Figure~\ref{fig:scanqa_sample}.

\begin{itemize}
    \item \textbf{Place}: \\
    Question: Where is the couch located? \\
    Answer: behind tables/ near small table/ against wall/ behind small round tables
    \item \textbf{Number}: \\
    Question: How many wooden tables are on either side of the blue couch? \\
    Answer: 3
    \item \textbf{Color}: \\
    Question: What color is the couch? \\
    Answer: blue/ dark blue
    \item \textbf{Object nature}: \\
    Question: What type of chair is next to the round brown table? \\
    Answer: dark blue couch / red ottoman chair \\
    Question: What kind of table is right behind a long black chair? \\
    Answer: round table / light brown round table \\
    Question: What shape is the wooden table? \\
    Answer: round/ circular round shape
    \item \textbf{Object}:\\
    Question: What is the red squared chair under? \\
    Answer: table/ round table in middle
\end{itemize}

\subsection{SQA3D}

SQA3D~\cite{sqa3d} is a benchmark that tests scene understanding of embodied agents. Given a 3D scene and a description of its position, the agent must first understand its situation and then reason and answer the question. The benchmark examines a wide spectrum of reasoning capabilities for an intelligent agent, ranging from spatial relation comprehension to commonsense understanding, navigation, and multi-hop reasoning, with examples in Figure~\ref{fig:sqa3dexample}.

\newpage

\section{Human-Intuition-Selection Method For Oracle View Evaluation}
\subsection{Method Introduction}

\label{sec:human-intuition-selection}
\begin{figure}[h!]
    \includegraphics[width=0.45\textwidth]{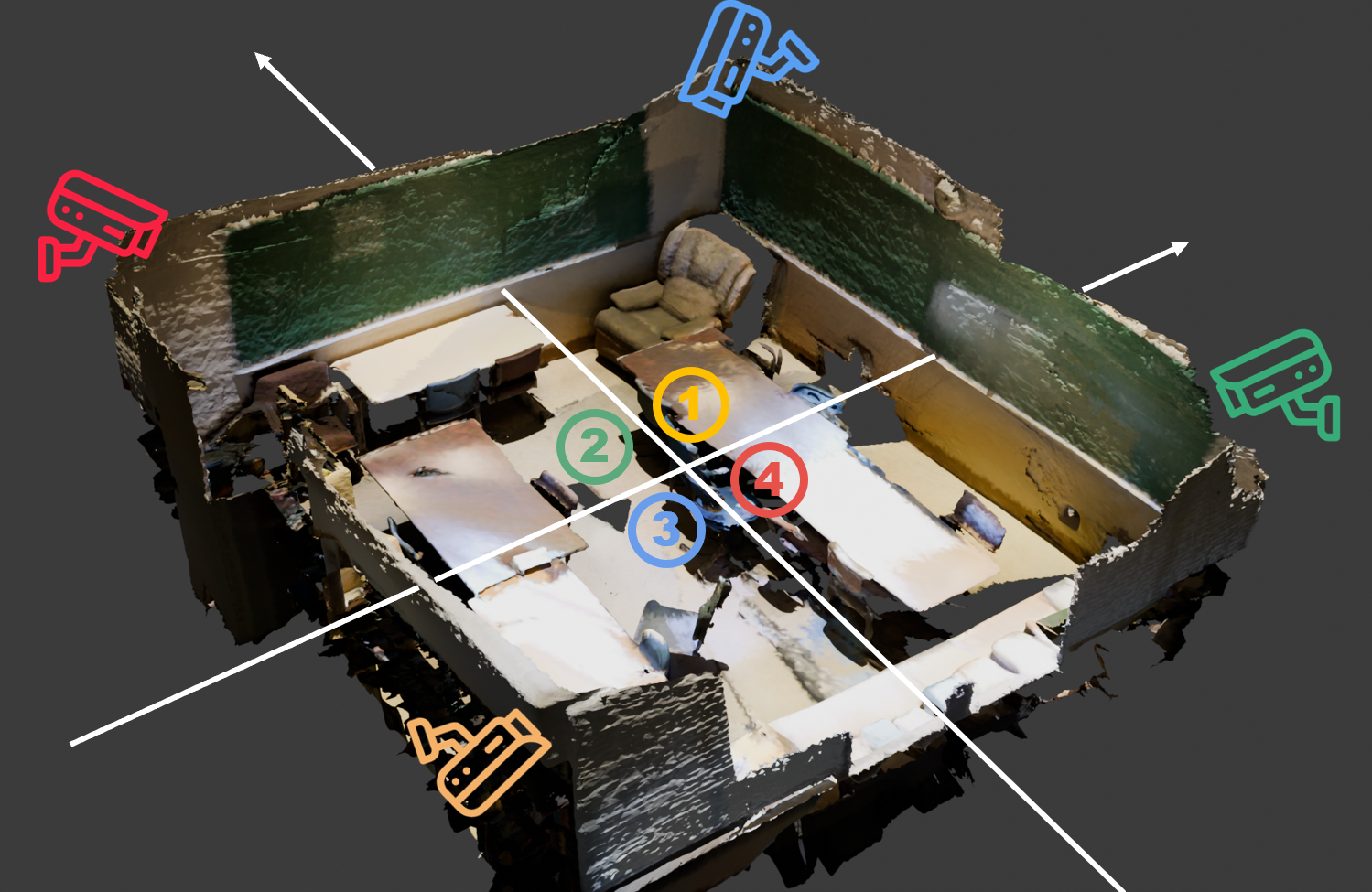}
    \caption{Illustration of the Human-Intuition-Selection Oracle Viewpoint Method: When the centroid of all relevant objects falls within a colored region, the viewpoint corresponding to the same color camera is selected as the oracle view.}
    \label{fig: HIS}
\end{figure}

ScanQA provides the \texttt{object\_id} in ScanNet~\cite{scannet} for all relevant objects of each question. Based on this information, the HIS method retrieves the bounding box and center coordinates for each relevant object and calculates the centroid of all relevant objects. As shown in Figure \ref{fig: HIS}, the scene is divided into four regions based on the camera angles (East, South, West, and North) for viewpoint number $k=4$. When the overall centroid falls within a specific region, the opposite viewpoint is selected as the oracle view. For viewpoint number $k=5$, we introduce an additional central region, and if the centroid falls within this area, the BEV viewpoint is chosen as the oracle view.

\subsection{Experimental Result}

\begin{table}[h!]
\centering
\small
\setlength{\tabcolsep}{4pt}
\resizebox{0.45\textwidth}{!}{
\begin{tabular}{lccc}
\toprule 
\textbf{Metric} & \textbf{Single View}& \textbf{Oracle View (HIS)}& \textbf{Oracle View (BoN)}\\
\midrule 
METEOR & 12.8 & 13.2 & 30.8 \\
ROUGH & 27.9 & 29.1 & 39.7  \\
CIDEr & 51.2 & 53.9 & 77.7  \\
\bottomrule 
\end{tabular}
}
\caption{Results of Different Oracle View Methods on ScanQA(val). HIS refers to Human-Intuition-Selection.}
\label{tab:two-oracle-view}
\end{table}

\newpage
\section{Case Study}
\label{sec: case study}

\subsection{Limitations of Text Similarity Metrics}
\label{sec: eval metric}

Using text similarity metrics for evaluation is too strict and often struggles to fully capture the nuances of model responses due to the openness and flexibility of natural language. Following are some examples from SQA3D and ScanQA, which use different text similarity metrics for evaluation:

\begin{itemize}
\item Task1 (From SQA3D)
\begin{itemize}
    \item Question: ``\textit{Is the table in front or at my back?}''
    \item Ground truth: ``\textit{back}''
    \item VLM3D's answer:  ``\textit{at my back}''
    \item Score: 0 (EM)
\end{itemize}
\item Task2 (From ScanQA)
\begin{itemize}
    \item Question: ``\textit{What shape is the window behind the table?}''
    \item Ground truth: ``\textit{rectangular window}'', ``\textit{rectangular}''
    \item VLM3D's answer:  ``\textit{rectangle}''
    \item Score: 0 (BLUE\_1), 0 (METEOR), 0 (CIDEr), 0 (ROUGE\_L)
\end{itemize}

\end{itemize}

Besides, benchmarks like ObjaverseXL-LVIS and Scan2Cap also employ text similarity metrics that rigidly evaluate syntactic patterns rather than semantic correctness. Such metrics fail to accommodate the open-domain nature of LLM responses, particularly for VLMs untrained on task-specific datasets to align with prescribed response styles.

\subsection{Ground Truth Preference}
\label{sec: ground truth prefer}

\subsubsection{3D MM-Vet}

\begin{figure}[h!]
    \includegraphics[width=\columnwidth]{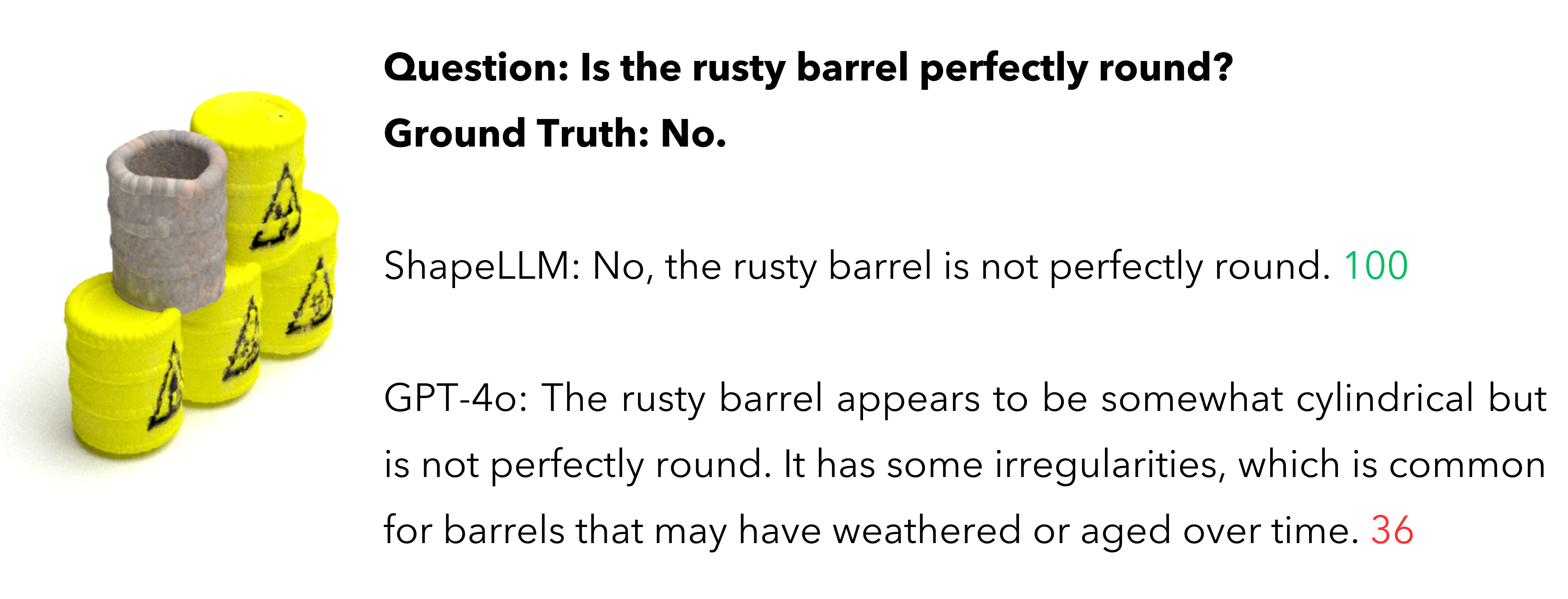}
    \caption{A case of ground truth preference from 3D MM-Vet.}
    \label{fig:3dmmvetcase}
\end{figure}

Figure~\ref{fig:3dmmvetcase} presents an example from 3D MM-Vet, where the task is to determine if a rusty barrel is perfectly round. The correct answer is ``No'', as the barrel is not perfectly round. Both ShapeLLM and GPT-4o reached similar conclusions, with ShapeLLM providing a brief answer that aligns more closely with the ground truth. However, VLM3D(GPT-4o) offered a more detailed explanation, resulting in a significantly lower score. It’s worth noting that this benchmark already uses large models to evaluate the answers.

\subsubsection{ScanQA}
\label{sec: ground truth prefer scanqa}
\begin{figure}[htbp]
  \begin{subfigure}[b]{\columnwidth}
    \includegraphics[width=\columnwidth]{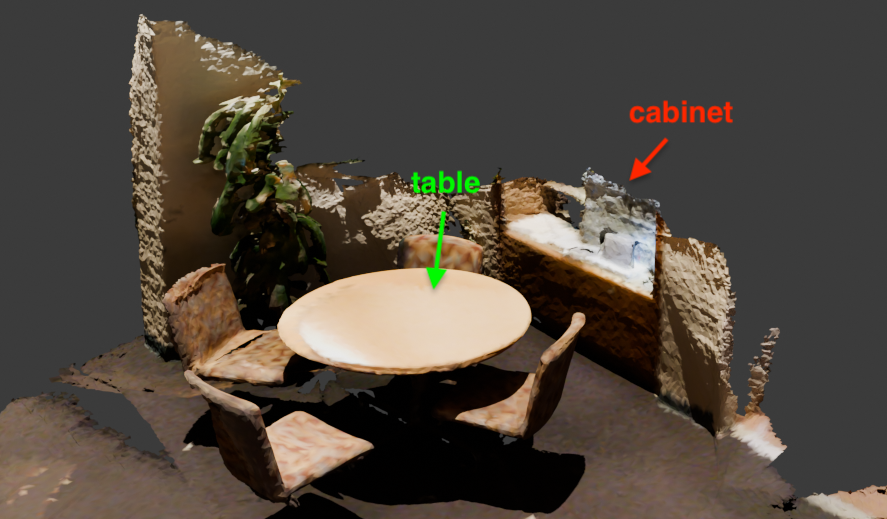}
    \subcaption{The viewpoint selected by Human-Intuition-Selection.}
  \end{subfigure}
  
  \hspace{2cm} 
  
  \begin{subfigure}[b]{\columnwidth}
  \includegraphics[width=\columnwidth]{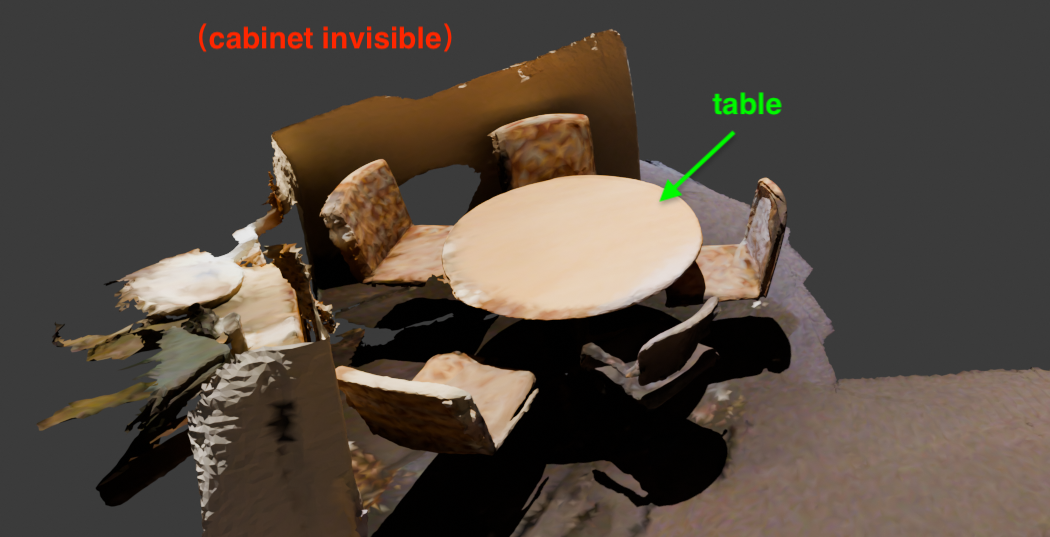}
  \subcaption{The viewpoint selected by Best-of-N.}
  \end{subfigure}
  \hspace{2cm} 
  \caption{Images used for each oracle view method. To better present the details, some blank areas in the images have been cropped.}
  \label{fig:gt preference}
\end{figure}

Here is an example where the Best-of-N method achieved a significantly better score due to its alignment with ground truth preference: the question is \textit{``Where is the brown chair located?''} and the ground truths are \textit{``under the circular table''} and \textit{``next to the left table''}. The HIS method gives the answer \textit{``near the cabinet''}, using the input image shown in Figure~\ref{fig:gt preference} (a). The answer provided by the Best-of-N method is \textit{``on the right side of the table''}, based on the input image shown in Figure~\ref{fig:gt preference} (b). 

It is evident that there are many reasonable answers to this question; however, the ground truth prefers responses related to the \textbf{table}. While both viewpoints depict the chair clearly, the answer from the Human-Intuition-Selection method is influenced by the nearby \textbf{cabinet} in its chosen viewpoint, resulting in a lower score.

\subsection{General Knowledge Over Specific Details}
\label{sec: general knowledge}
\begin{figure}[h!]
    \includegraphics[width=\columnwidth]{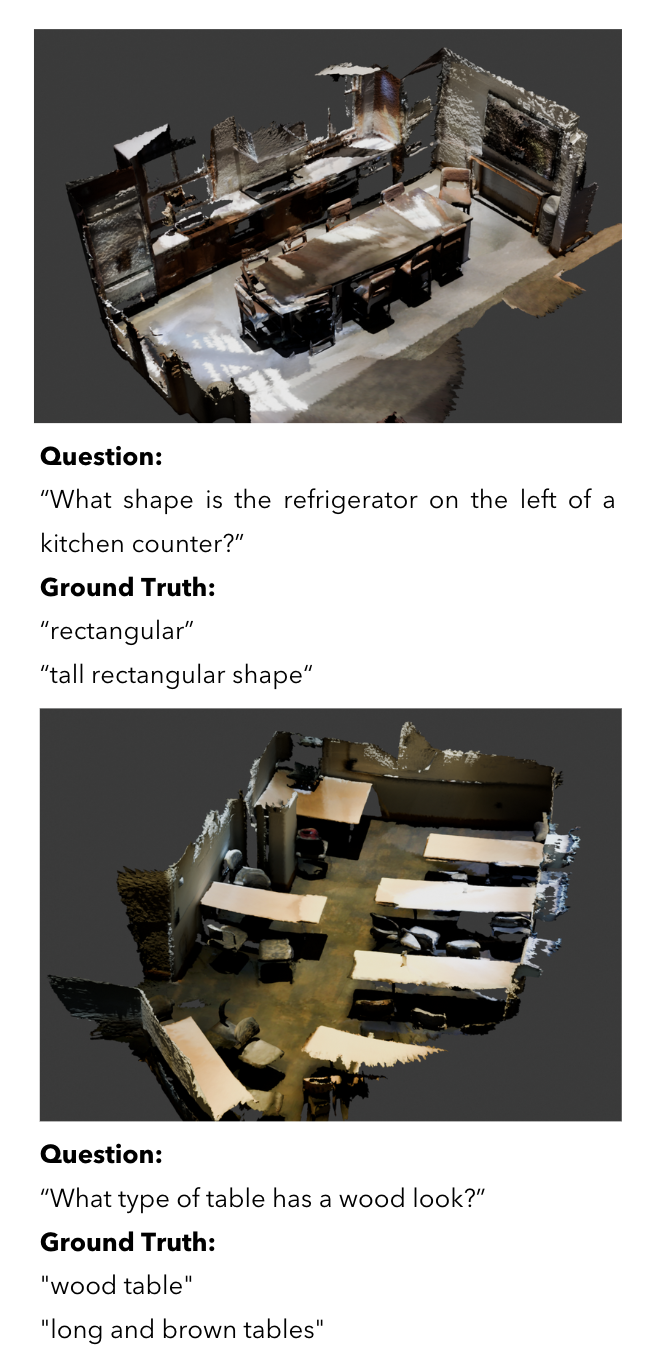}
    \caption{Example tasks from ScanQA.}
    \label{fig: world knowledge}
\end{figure}

Figure~\ref{fig: world knowledge} presents several examples from the case study. It is evident that the questions can be resolved through general world knowledge, without relying on the unique characteristics of the scenes involved.

\begin{figure}[htbp]
  \begin{subfigure}[b]{\columnwidth}
    \includegraphics[width=\columnwidth]{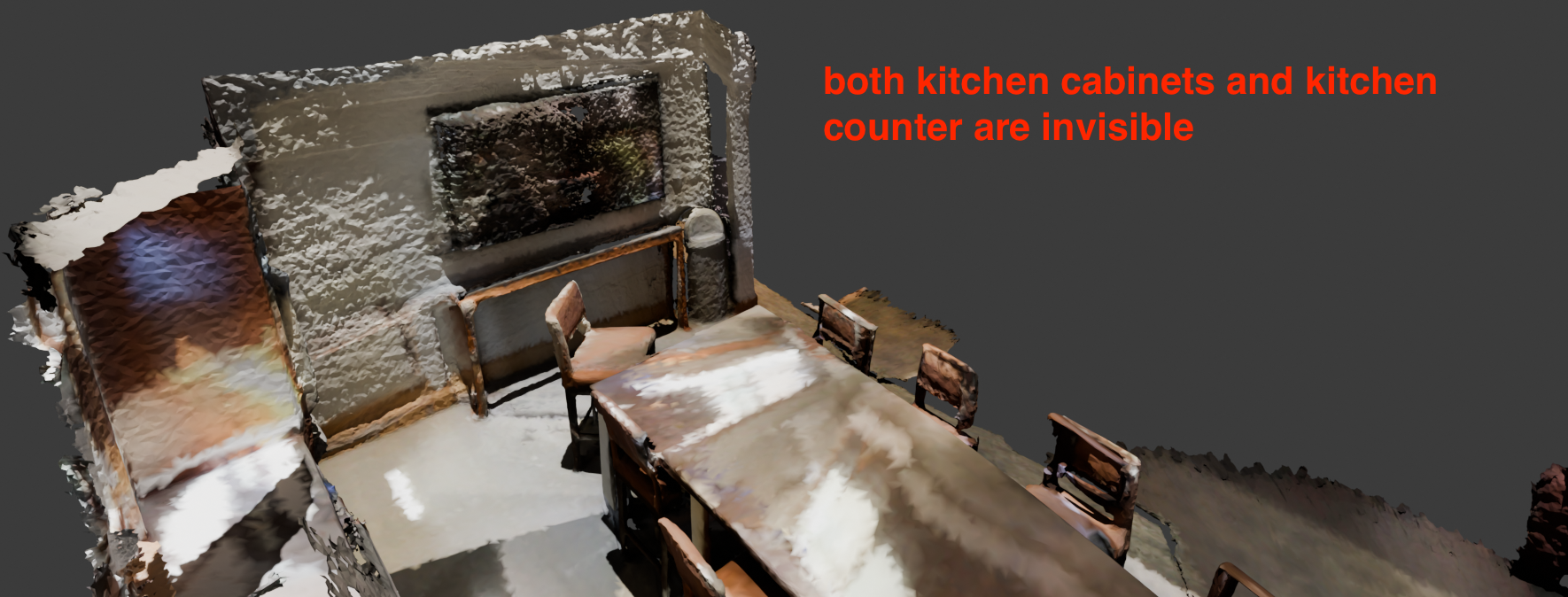}
    \subcaption{The viewpoint selected by Best-of-N.}
  \end{subfigure}
  \hspace{2cm} 

   \begin{subfigure}[b]{\columnwidth}
    \includegraphics[width=\columnwidth]{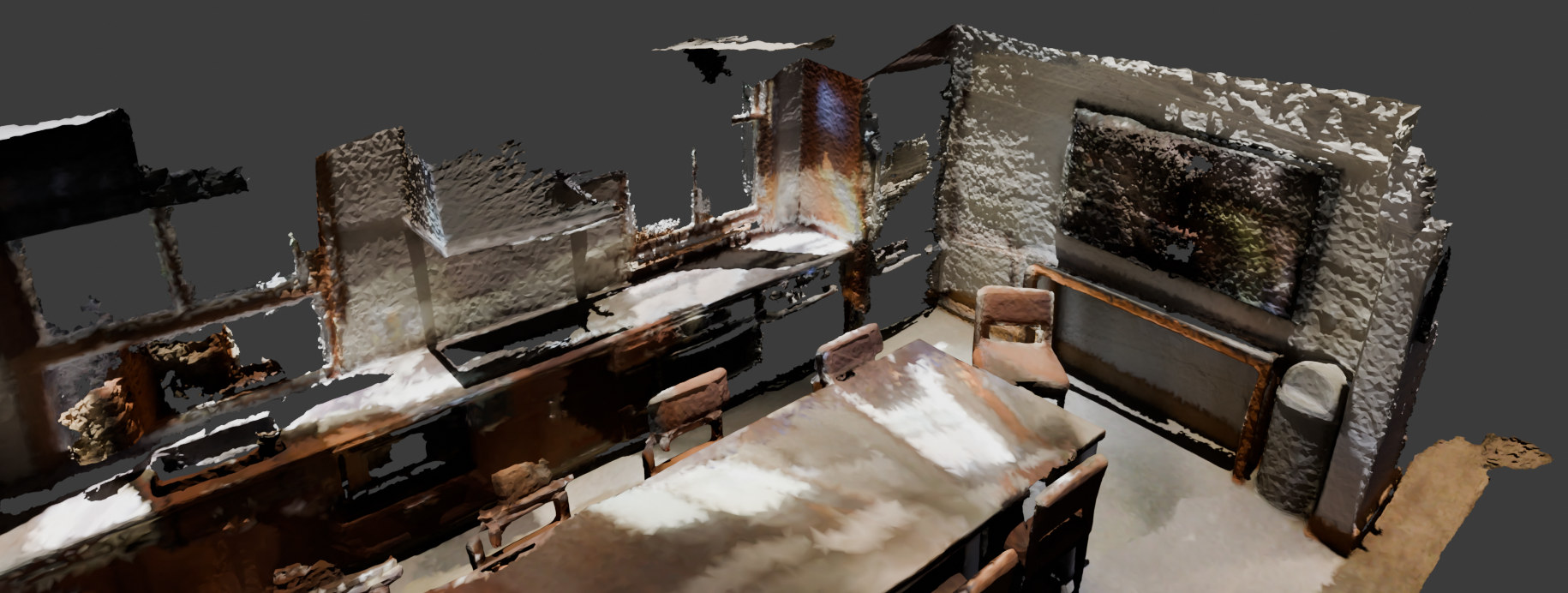}
    \subcaption{The viewpoint selected by Human-Intuition-Selection.}
  \end{subfigure}
  \hspace{2cm} 

  \caption{Images used for each oracle view method. To better present the details, some blank areas in the images have been cropped.}
  \label{fig: world knowledge 2}
\end{figure}

Besides, here is an example where the Best-of-N method answers the question using an irrelevant viewpoint and still achieves better results. The reason is that the question is too general and can be answered solely based on world knowledge; however, the Human-Intuition-Selection method provides a more specific answer. The question is \textit{``What is under the long kitchen counter?''} and the ground truths are \textit{``brown rectangular kitchen cabinets''} and \textit{``brown kitchen cabinets''}. The answer provided by the Best-of-N method is \textit{``kitchen cabinet''}, based on the input image shown in Figure~\ref{fig: world knowledge 2} (a). The Human-Intuition-Selection method gives the answer \textit{``black dishwasher''}, using the input image shown in Figure~\ref{fig: world knowledge 2} (b).

In the image used by the Best-of-N method, neither the kitchen counter nor kitchen cabinets are explicitly shown. However, in a typical kitchen layout, the area beneath the kitchen counter is commonly referred to as 'kitchen cabinets'. This can be answered using general world knowledge, which allows the method to successfully address the vague question. In contrast, the Human-Intuition-Selection method employs a viewpoint that clearly presents the objects related to the question, providing a more specific answer: ‘black dishwasher’. However, the question is intended to assess the general area beneath the counter as a whole, which results in this more specific answer receiving a lower score.

\newpage
\section{Experimental Details}
\subsection{Model Size}

We use Qwen2-VL with 72B parameters in all experiments.

\subsection{Rendering}
\label{sec:experimental_details}

For the two object point cloud benchmarks, we used the Mitsuba renderer due to its efficiency and lightweight nature. For the two scene point cloud benchmarks, which utilize mesh-based point clouds, we used the Blender rendering engine, consistent with the methodology used in the original work \cite{scanqa, sqa3d}.

We optimized the rendered images by adjusting key parameters such as lighting configurations and point cloud particle size. For more specific details, please refer to our code repository.

\subsection{Evaluation}
We replaced the GPT-3.5 model originally used for evaluation in the 3D-MM-Vet benchmark with GPT-4o-mini.

\end{document}